\documentclass[twoside]{article}

\usepackage[round]{natbib} % omit 'round' option if you prefer square brackets
\usepackage{graphicx}
\usepackage{amssymb}
\usepackage{hyperref}
\usepackage{bbm}
\usepackage{caption}
\usepackage{subcaption}

\usepackage[accepted]{aistats2023}
\usepackage[round]{natbib}

\newtheorem{assumption}{Assumption}

\begin{document}


\runningauthor{Chauhan, Molaei, Tania, Thakur, Zhu and Clifton}

\twocolumn[

\aistatstitle{Adversarial De-confounding in Individualised Treatment Effects Estimation}

\aistatsauthor{Vinod Kumar Chauhan$^1$ \And Soheila Molaei$^1$ \And  Marzia Hoque Tania$^1$}

\aistatsauthor{Anshul Thakur$^1$ \and Tingting Zhu$^1$ \and David A. Clifton$^{1,2}$}

\aistatsaddress{$^1$Institute of Biomedical Engineering, University of Oxford, UK\\$^2$Oxford-Suzhou Centre for Advanced Research, Suzhou, China}
]

\begin{abstract}
    Observational studies have recently received significant attention from the machine learning community due to the increasingly available non-experimental observational data and the limitations of the experimental studies, such as considerable cost, impracticality, small and less representative sample sizes, etc.
    In observational studies, de-confounding is a fundamental problem of individualised treatment effects (ITE) estimation.
    This paper proposes disentangled representations with adversarial training to selectively balance the confounders in the binary treatment setting for the ITE estimation. The adversarial training of treatment policy selectively encourages treatment-agnostic balanced representations for the confounders and helps to estimate the ITE in the observational studies via counterfactual inference.
    Empirical results on synthetic and real-world datasets, with varying degrees of confounding, prove that our proposed approach improves the state-of-the-art methods in achieving lower error in the ITE estimation.
\end{abstract}

\section{Introduction}
\label{sec_intro}
Individualised treatment effects (ITE) estimation is a fundamental problem that is useful for making personalised decisions and estimating their effects. For example, in the intensive care unit (ICU), ITE can be used to decide whether or not to give a medication to a patient, which can be a question of life or death of the patient. The ITE estimation learning requires answering counterfactual questions, such as: ``\textit{What would have been the outcome if alternative treatment had been given?}", i.e., it requires predicting potential outcomes of unexplored actions (\cite{rubin2005causal}). Due to its importance, ITE estimation is studied widely across diverse fields, like medicine, marketing, education and policy-making, etc. (please refer to \cite{bica2021real} for an overview).

Randomised controlled trial (RCT) experiments are the gold standard to evaluate the effectiveness of treatments (\cite{pearl2009causality}). However, they are expensive, time-consuming, sometimes unethical and impractical, and have small and less representative sample sizes, etc. On the other hand, non-experimental observational studies are becoming popular for evaluating the effectiveness of treatments due to increasingly available observational data, like electronic health records, and overcoming limitations of the RCT.

The ITE estimation from the observational studies have, recently, received a great attention from the machine learning community, e.g., \cite{shalit2017estimating,shi2019adapting,hassanpour2019learning,curth2021inductive,curth2021nonparametric,wu2022learning} etc., and it is different from standard machine learning (\cite{rubin2005causal,pearl2009causality,wager2018estimation}). This is because the observational data have outcomes available only for the actions taken, i.e., for the selected treatments (factual outcomes), but the outcomes for alternative treatments are not available (counterfactual outcomes) -- which is called \textit{the fundamental problem of causal learning} (\cite{holland1986statistics}). Moreover, the observational data contain confounders, i.e., covariates which affect outcomes as well as treatment assignment policy, and hence have selection-bias (\cite{imbens2015causal}) (i.e., $p(T=0 \vert X=x) \neq p(T=1 \vert X=x)$, unlike the RCT where treatments are assigned randomly and has $p(T=0 \vert X=x) = p(T=1 \vert X=x)$, where $p(T=t\vert X=x)$ is probability of treatment $t$ for a given patient $x$ in a binary treatment setting). For example, in the above ICU setting scenario, suppose choice of medication is dependent on the age of the patient which also affects the recovery rate.

ITE estimation from observational data is a challenging problem and it involves a fundamental problem of \textit{de-confounding}, i.e., removing selection-bias which otherwise precludes direct comparison between different treatment groups as they have different distributions. Without careful handling of selection-bias, it could lead to biased estimates of causal effects (\cite{zubizarreta2015stable}). Recently, several deep learning architectures and algorithms were proposed to address the de-confounding, which borrow ideas from representation learning (\cite{bengio2013representation}), domain adaptation (\cite{johansson2016learning}) and disentanglement learning (\cite{hassanpour2019learning}).

\begin{figure}[htb!]
    \centering
    \includegraphics[width=0.35\textwidth]{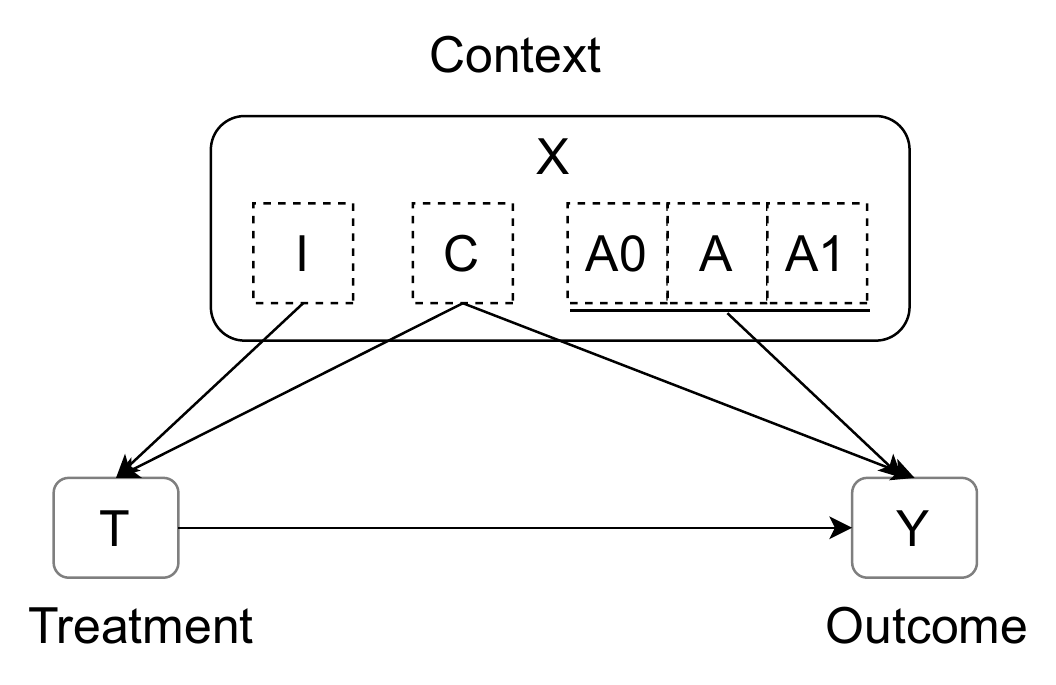}
    \caption{Problem setting: We consider binary treatment setting where the context $X$ can be decomposed into five (or three) subsets as covariates $I$ which affect only the treatment policy $\pi$, named instrument variables, covariates $\lbrace A0, A, A1 \rbrace$ which affect only the treatment outcomes $Y$, where $A0$ affects only the control group outcomes $\mu_0$, $A1$ affects only the treatment group outcomes $\mu_1$, and $A$ affects both of the outcomes, collectively referred to as adjustment variables and covariates $C$ which affect both treatment policy as well as treatment outcomes.}
    \label{fig_problem_setting}
\end{figure}

Representation learning-based methods try to reduce selection-bias by learning a shared representation for outcome and treatment predictions (\cite{johansson2016learning}). Similarly, domain adaption-based methods try to learn a common representation space and minimise the discrepancy between control and treatment distributions (\cite{shalit2017estimating}). Disentanglement learning-based methods decompose the covariates into latent factors corresponding to confounders (i.e., features affecting both treatment and outcomes), instrument variables (i.e., features affecting only treatment decision) and adjustment variables (i.e., features affecting only outcome predictions) for counterfactual interference (\cite{hassanpour2019learning}), as shown in Fig.~\ref{fig_problem_setting} which presents the problem setting.

This paper proposes novel disentangled representations with adversarial training to selectively balance the confounders in the binary treatment setting for ITE estimation. The disentanglement representation learning -- using a specific deep learning architecture and constraining orthogonality on each pair of representations -- helps to disentangle the latent factors, i.e., representations for the confounders, instrument variables and adjustment variables (refer to Fig.\ref{fig_architecture}). The adversarial training jointly optimises the confounder representations to minimise the loss for the potential outcome predictions and maximise the loss for the treatment assignments. This learns a representation for the confounders, which is treatment-agnostic. Our empirical results on synthetic and real-world datasets, with varying degrees of confounding, show that our proposed approach improves the state-of-the-art methods in achieving lower error in estimating the ITE.

The rest of the paper is organized as: Section~\ref{sec_literature} presents some related work, Section~\ref{sec_problem} defines the problem settings, Section~\ref{sec_methodology} describes the proposed methodology and Section~\ref{sec_experiments} presents experiments. Finally, the concluding remarks and future scope are discussed in Section~\ref{sec_conclusion}.

\section{Related Work}
\label{sec_literature}
The estimation of individualised treatment effects (ITE) from the observational data has recently received growing attention from the machine learning community due to the increasingly available observational data, like electronic health records, and overcoming limitations of randomised controlled trials -- the gold standard to estimate treatment effects. ITE estimation from the observational data faces a fundamental problem of de-confounding and the corresponding selection-bias. As discussed below, several techniques have been proposed for de-confounding, such as weighting of outcomes, learning balanced representations, and dis-entanglement learning.

\textbf{Weighting techniques} to address the confounders typically construct some weights to balance the covariates for the control and the treatment groups (\cite{rosenbaum1983central}), e.g., inverse probability weighting (IPW). In reality, it is difficult to estimate the correct propensity score values, i.e., the probability of treatment, so these methods are known to be unstable (\cite{hainmueller2012entropy}). To address this limitation, several alternative weighting schemes are proposed, such as truncated IPW (\cite{crump2009dealing}), matching weights (\cite{li2018balancing}) and calculating weights using optimisation program (\cite{li2017matching}) etc. The weighting techniques are known not to scale well to high dimensional problems.

\textbf{Balanced representation techniques} learn a shared representation, i.e., an embedding for the confounders, and force the distributions of the control and treatment arms to be similar. For example, \cite{johansson2016learning,shalit2017estimating} presented the pioneering work and borrowed the idea of discrepancy from the domain adaptation to introduce a form of regularisation to enforce similarity in the representations for control and treatment groups. Rather than balancing the global distributions for different treatment groups, \cite{yao2018representation} extended the idea of representation learning by local similarity preserving technique. The combination of weighting and representation techniques is also explored in \cite{hassanpour2019counterfactual}, where authors estimated IPW from the representations to re-weight the regression terms.

Like weighting techniques, representation learning techniques focus on the confounders and balance the representations for the whole context. Recently, \cite{vanderweele2019principles,kuang2019treatment} highlighted the importance of confounder separation for causal learning and argued that by balancing the complete context, some variables, such as instrument variables, lead to additional bias and variance.

\textbf{Disentanglement representation techniques} address the above concerns of balanced representation learning techniques. They learn to disentangle the context $X$ into latent factors for subsets of $X$. For example, \cite{hassanpour2019learning} proposed disentanglement learning algorithm to learn three latent factors $\lbrace C(X), I(X), A(X) \rbrace$ for confounders $C$, instrument variables $I$ and adjustment variables $A$ from the context $X$. However, they didn't guarantee the disentangled representations and used propensity scores for weighting, which runs the risk of instability similar to weighting techniques. The idea of disentanglement is further improved by different researchers and is currently the state-of-the-art technique for ITE estimation. For example, \cite{wu2022learning} addressed the above limitations, removed the propensity scores, and introduced orthogonality between each pair of representations, ensuring that each variable in the context $X$ affects only one of the three representations, i.e., $\lbrace I, C, A\rbrace$. Recently, \cite{curth2021nonparametric} proposed to disentangle the context $X$ into five latent factors $\lbrace C(X), I(X), A0(X), A(X), A1(X) \rbrace$, corresponding to confounders, instrument variables, adjustment variables, where $A0, A, A1$ refer to adjustment variables which affect the control group, control as well as treatment groups and treatment group, respectively, and used orthogonality to enforce that each variable in $X$ affects only one of these five factors.

In this paper, we proposed a novel disentangled representation with adversarial training to selectively balance the confounders. The disentanglement learning helps to separate the confounds and adversarial training (\cite{ganin2016domain}) selectively learns balanced representations for the confounders, unlike the balanced representation learning methods which balance the whole context $X$ and may lead to additional bias and variance in counterfactual inference.

\section{Problem Setting}
\label{sec_problem}
We introduce the problem of ITE estimation for binary treatments from the observational data using the potential outcomes framework (\cite{rubin2005causal}). Suppose, $D = \lbrace T_i, X_i, Y_i \rbrace_{i=1}^m$ is a sample of $m$ units, say patients, of the observational data, e.g., electronic health records, taken \textit{i.i.d.} from an unknown distribution $\mathbb{P}$. $X_i \in \mathbb{R}^d$ is a $d$-dimensional covariates, i.e., context, e.g., patient history. $T_i \in \lbrace 0, 1 \rbrace$ is a binary treatment variable, where $T_i=1$ refers to a patient $i$ receiving a treatment and $T_i=0$ refers to a patient not receiving the treatment. $Y_i \in \mathbb{R}$ denotes the patient's outcome.

We use Neyman-Rubin potential outcomes framework (\cite{rubin2005causal}) for the individualised treatment effects.
% , which is defined as the difference between potential outcomes, i.e., 
% \begin{equation}
%     \label{eq_ite}
%   ITE_i = Y_i(1) - Y_i(0),
% \end{equation}
Suppose $Y_i(1)$ and $Y_i(1)$ refer to potential outcomes when patient $i$ receives a treatment, i.e., $T_i=1$ and did not receive the treatment, i.e., $T_i=0$, respectively. However, due to \textit{the fundamental problem of the causal inference}, we observe only factuals, i.e., the outcome for the selected treatment for a given patient, and not the counterfactuals, i.e., the outcome for the non-selected treatment, as the outcome of the patient is defined as $Y_i = T_i Y_i(1) - (1-T_i) Y_i(0)$. The ITE is the conditional average treatment effect, also known as CATE\footnote{in machine learning literature, ITE and CATE are interchangeably used to refer to CATE, and is different from $ITE_i = Y_i(1) - Y_i(0)$ in causal inference literature (\cite{johansson2020generalization})}, as given below.
\begin{equation}
    \label{eq_cate}
    e(x) = \mathbb{E}_\mathbb{P} \left[ Y(1) - Y(0) \vert X=x \right].
\end{equation}

Our work is based on standard assumptions of treatment effects estimation (\cite{imbens2015causal}), which are given below.
\begin{assumption}
\label{sutva}
    (\textbf{Stable Unit Treatment Value}:) The distribution of a patient's outcome depends on the treatment given to it and is independent of the treatment given to other patients.
\end{assumption}

\begin{assumption}
\label{unconfoundedness}
    (\textbf{Ignorability}: \cite{rosenbaum1983central}) The treatment assignment policy is independent of the potential outcomes when given the context of a patient, 
    \begin{equation}
    \label{eq_unconfoundedness}
        \lbrace Y(0), Y(1) \rbrace~\perp~T \vert X.
    \end{equation}
\end{assumption}
Assumption~\ref{unconfoundedness} is also known as unconfoundedness because it holds if there are no hidden confounders and if these satisfy additional conditions.
\begin{assumption}
\label{overlap}
    (\textbf{Overlap}: \cite{imbens2004nonparametric}) The treatment assignment policy is stochastic, i.e., each patient has a certain probability of receiving any treatment, i.e.,
    \begin{equation}
    \label{eq_overlap}
        0 < \pi(x) < 1, \quad \forall x \in X,
    \end{equation}
    where $\pi(x)$ is probability of receiving treatment $T=1$ of patient $x$.
\end{assumption}
The Assumptions~\ref{unconfoundedness} and \ref{overlap} together are called \textit{strong ignorability} assumptions (\cite{imbens2009recent}).

Under the above assumptions, expected potential outcomes and the conditional average treatment effect are identifiable as given below.
\begin{equation}
    \label{eq_po}
    \mu_t(x) = \mathbb{E}_\mathbb{P} \left[ Y \vert X=x, T=t \right],
\end{equation}
\begin{equation}
\label{eq_cate2}
    \begin{split}
        e(x) & = \mathbb{E}_\mathbb{P} \left[ Y(1) - Y(0) \vert X=x \right],\\
        & = \mathbb{E}_\mathbb{P} \left[ Y \vert X=x, T=1 \right] - \mathbb{E}_\mathbb{P} \left[ Y \vert X=x, T=0 \right],\\
        & = \mu_1(x) - \mu_0(x).
    \end{split}
\end{equation}

It is well-known that the above assumptions are fundamentally untestable from given observational data for ITE estimation (\cite{pearl2009causality}), so domain knowledge is necessary to access the validity of these assumptions (\cite{bica2021real}). Moreover, richer datasets can also cover more of the confounding variables.

\section{Methodology}
\label{sec_methodology}
The individualised treatment effects can be estimated from the observational data, $D = \lbrace T_i, X_i, Y_i \rbrace_{i=1}^m$, containing context $X_i \in \mathbb{R}^d$, treatment assignments $T_i \in \lbrace 0, 1\rbrace$ and outcomes $Y_i \in \mathbb{R}$ for the patients $i=1,2,...,m$ by training supervised machine learning model for potential outcome regressions $\mu_t(x) = \mathbb{E}_\mathbb{P} \left[ Y \vert X=x, T=t \right]$ and using plug-in approach as $e(x) = \mu_1(x) - \mu_0(x)$. However, it needs to address the confounders (i.e., features affecting the treatment policy as well as treatment outcomes) and the resulting selection-bias in the treatment assignment policy (\cite{pearl2009causality}). To address the selection-bias caused by the confounders, we separate the confounders using the disentanglement representation learning (\cite{hassanpour2019learning,curth2021nonparametric,wu2022learning}) and selectively learn balanced confounder representations using the adversarial training (\cite{ganin2016domain}) of expected outcome predictions against the propensity score (i.e., probability of treatment) predictions.

\subsection{Disentangled Representations}
\label{subsec_disentanglement}

Suppose the context $X$ can be decomposed into five latent factors $\lbrace I(X), C(X), A0(X), A(X), A1(X) \rbrace$ for instrument variables $I$ (variables which affect only the treatment policy), adjustment variables $A, A0$ and $A1$ that affect both the potential outcome regressions $\mu_0$ and $\mu_1$, only $\mu_0$ and only $\mu_1$, respectively, and confounder variables $C$ (variables affecting the treatment policy as well as the outcomes). To implement the disentanglement representation learning, it requires a deep learning architecture that has five multi-layer perceptrons (MLP) and takes context $X$ as an input to generate latent representations, i.e., embedding for each of the five factors, $I, A, A0, A1$ and $C$, as well as orthogonality \cite{curth2021nonparametric,wu2022learning} between each pair of the representations to ensure each variable in $X$ contributes to only one of the five factors. We present the deep learning architecture to support disentanglement in Fig.~\ref{fig_architecture} and the idea of orthogonality is given below.
\begin{figure}[htb!]
    \centering
    \includegraphics[width=0.5\textwidth]{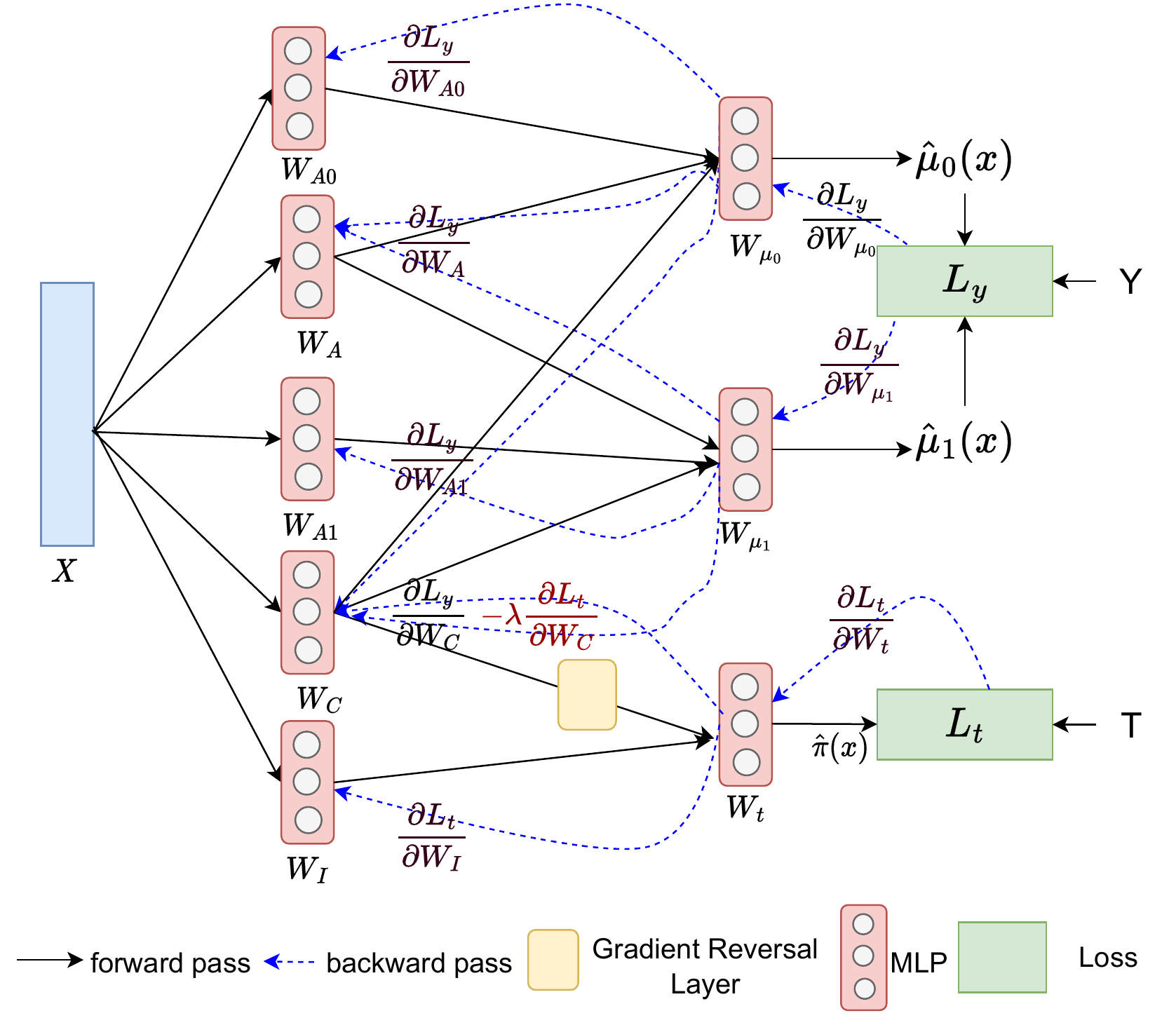}
    \caption{SNet+: proposed disentanglement representation based adversarially balanced architecture. The context $X$ is disentangled into five latent factors $\lbrace I(X), C(X), A0(X), A1(X), A(X)\rbrace$. The treatment classifier is adversarially trained by using a gradient reversal layer between confounder representations $C(X)$ and the treatment classifier to get balanced representations for the confounders.}
    \label{fig_architecture}
\end{figure}

The architecture-based representations of $\lbrace I, A, A0, A1, C \rbrace$ are insufficient to guarantee disentanglement as data-driven neural networks overfit the training. So, similar to \cite{kuang2017treatment,curth2021nonparametric,wu2022learning}, we use orthogonality to enforce variable decompositions among the five representation networks $\lbrace I(X), A(X), A0(X), A1(X), C(X) \rbrace$ corresponding to $\lbrace I, A, A0, A1, C \rbrace$, respectively. Let's take an example of a representation network for confounders $C(X)$ and suppose it takes $l$ layers and weights of the $k$th layer are denoted by matrix $W_k$. Then, the contribution of each variable in $X$ on each dimension of representation $C(X)$ can be approximated by computing $W_1 \times W_2 \times ... \times W_l$ which can be denoted as $W_C \in \mathbb{R}^{d\times n}$, where $d$ is the dimension of $X$ and $n$ is the dimension of $C(X)$. We take the average along the rows of $W_C$ to represent the average contribution of variables in $X$ on representations $C(X)$, denoted as $\Bar{W_C}$. Similarly, we approximate the contributions of each variable in $X$ on representations of $I(X), A(X), A0(X), A1(X)$ as $\Bar{W_I}, \Bar{W_A}, \Bar{W_{A0}}, \Bar{W_{A1}}$, respectively. Assuming all representation networks have the same structure, the hard decomposition can be enforced by constraining orthogonality on each pair of them. It can be achieved by adding the following loss term to the objective function.
\begin{equation}
    \label{eq_orthogonolisation}
    \begin{split}
        L_O & = \Bar{W_I}^T.\Bar{W_A} + \Bar{W_I}^T.\Bar{W_{A0}} + \Bar{W_I}^T.\Bar{W_{A1}} + \Bar{W_I}^T.\Bar{W_C} \\
        &+ \Bar{W_A}^T.\Bar{W_{A0}} + \Bar{W_A}^T.\Bar{W_{A1}} + \Bar{W_A}^T.\Bar{W_C} + \Bar{W_{A0}}^T.\Bar{W_{A1}} \\
        & + \Bar{W_{A0}}^T.\Bar{W_C} + \Bar{W_{A1}}^T.\Bar{W_C}.
    \end{split}
\end{equation}
This loss term can lead to the result $\Bar{W}^k_I = \Bar{W}^k_A = \Bar{W}^k_{A0} = \Bar{W}^k_{A1} = \Bar{W}^k_C = 0$. So, to avoid this, following regulariser as a soft constraint is used, which make the sum of each of $\Bar{W_I}, \Bar{W_A}, \Bar{W_{A0}}, \Bar{W_{A1}}, \Bar{W_{C}}$ to 1, as given below.
\begin{equation}
    \label{eq_ortho_regulariser}
    R_O = \sum_{r \in \lbrace I, A, A0, A1, C \rbrace } \left(\sum_{i=1}^d \Bar{W}^i_r - 1\right)^2.
\end{equation}
The representation networks for the five latent factors, $\lbrace I(X), A(X), A0(X), A1(X), C(X) \rbrace$, and the above loss term and the regulariser help to learn disentangled representations for separating the confounders from the input context $X$. Similarly, the context can be decomposed into three latents $\lbrace I(X), A(X), C(X) \rbrace$ corresponding to $\lbrace I, A, C \rbrace$, respectively, and referred to as DRCFR+.

\subsection{Adversarial De-confounding}
\label{subsec_adversarial_deconf}
Domain adaptation is the task of building a predictor/classifier model on the source domain while being invariant to the shifts in the target domain, where the source and target domains have similar but different distributions. \cite{johansson2016learning,shalit2017estimating} presented pioneering work to formulate the ITE estimation problem as domain adaptation and learnt shared representations, i.e., embedding $\phi(X)$, and used integral probability metrics to minimise the distance between control and treatment distributions in the embedding space, thus leading to learning balanced representations to get rid off the selection-bias. However, later \cite{vanderweele2019principles,kuang2019treatment} pointed out that learning such balanced representations for the context $X$ can lead to additional bias and variance in the predictions.

Here, we adopt another popular idea of adversarial training (\cite{ganin2016domain}) from the domain adaptation which is widely adapted in different problem settings, e.g., \cite{bica2020estimating,guan2021domain}. The main idea of domain adaptation is to obtain invariance between source and target distributions. In our setting, we want invariance of potential outcome regressions $\mu_t$ against the treatment policy $\pi$, which is also different than the \cite{johansson2016learning,shalit2017estimating} who didn't consider treatment policy for reducing the selection-bias. Adversarial training helps to learn representations for the confounders such that the representations are discriminative of the potential outcome regressions $\mu_t$ but invariant to the treatment policy $\pi$. This results in balanced representations for control and treatment groups, which also avoids the risk of additional bias and variance, as pointed out in \cite{vanderweele2019principles,kuang2019treatment} from balancing the entire context $X$.

The balanced representations are achieved by jointly optimising the underlying representations as well as a predictor and a classifier on these representations: (i) potential outcome predictor which is used during the training and test times, and (ii) treatment policy classifier which is used during the training time only. The parameters of the predictor and classifier $\lbrace W_{\mu_0}, W_{\mu_1}, W_t \rbrace$ (refer to figure Fig.~\ref{fig_architecture}) are optimised to reduce the training error and parameters of the confounder representation $W_C$ are optimised to be better predictive of potential outcomes, i.e., to \textit{minimise} the potential outcome loss, and to be invariant to the treatment policy, i.e., \textit{maximise} the loss of the treatment classifier and hence push the treatments to be similar to the randomised controlled trials. Thus, the later update works \textit{adversarially} to the treatment classifier, and enforces treatment-invariant balanced representations to the confounders $C$.

To implement the adversarial training to get the balanced representations for the confounders, a gradient reversal layer is placed between the confounder representations $C(X)$ and the treatment classifier, as shown in Fig.~\ref{fig_architecture}. During the forward pass, the layer acts as an identity, but during the backward pass, the layer reverses the direction of the gradient, as the layer's name suggests. A domain adaptation hyperparameter $\lambda$ is used to control the balancing effect of the gradient reversal layer and is defined as follows.
\begin{equation}
    \label{eq_doamin_param}
    \lambda = \lambda_0 \left(\dfrac{2}{1 + exp(-10 \times epoch/\gamma)} -1\right),
\end{equation}
where $epoch$ is the epoch number, $\gamma$ controls the variations of $\lambda$ during the training, and it can take values from one to the total number of epochs. $\lambda$ can vary from 0 to a large number, which has to be tuned as per the dataset. The constant factor $\lambda_0$ is one unless we need $\lambda$ to be more than one. We observe that the estimator is sensitive to the choice of $\lambda$, which controls the balancing effect. Since synthetic and semi-synthetic datasets come with different variants which differ in their outcome surfaces, e.g., the IHDP dataset has 100 variations, it is difficult to find a common value for all the variants which affect the performance of the proposed estimators.

The objective function for the proposed ITE estimator is as given below.
\begin{equation}
\begin{split}
    \frac{1}{m}\sum^m_{i=1} \left[L_y(y_i,\hat{\mu}_t^i(x)) + L_t(t, \hat{\pi}(x))\right] + \lambda_1 \mathcal{R}_2 + \lambda_2 (L_O + R_O),
\end{split}
\end{equation}
where $L_y$ is loss for potential outcomes, $L_t$ is propensity score, $R_2$ is $l_2$-regulariser, and $\lambda_1$ and $\lambda_2$ are constants.

% The idea of domain adversarial training can be applied to a wide range of existing ITE estimators which learn confounder representations, as we will be used in the following section.

\section{Experiments}
\label{sec_experiments}
This section, discusses the experimental settings, metrics, baselines and results on synthetic and real datasets.

\subsection{Experimental Setting}
\label{subsec_setting}
Following \cite{curth2021nonparametric}, we fix common hyperparameters for all the ITE estimators. We take a mini-batch of 100, Adam as optimiser, the learning rate of 0.0001, and $l2$-regulariser coefficient of 0.0001. We use early stopping with patience of 50 as stopping criteria (for more details of the hyperparameters and implementation details, please refer to supplementary material). All the experiments are implemented in Pytorch, executed on an Ubuntu machine (64GB RAM, 1 NVIDIA GeForce GPU), and averaged over 10 runs. The final code will be released on GitHub. For baselines, we use the following ITE estimators.\\
\textbf{FlexTENet} (\cite{curth2021inductive}): It has a flexible architecture that allows flexibility to share information at different layers between potential outcomes for factual and counterfactuals. Moreover, the authors also introduced regularisation-based schemes to encourage similarity between the two outcomes.\\
\textbf{SNet} (\cite{curth2021nonparametric}): It decomposes the context $X$ into five latent factors and uses orthogonality to encourage each variable of $X$ to contribute to one of the five latent factors.\\
\textbf{DRCRF} (\cite{hassanpour2019counterfactual,wu2022learning}): It decomposes the context $X$ into three latent factors and uses orthogonality to encourage each variable of $X$ to contribute to one of the three latent factors.\\
\textbf{TARNet, CFRNet} (\cite{johansson2016learning,shalit2017estimating}): This pioneering work introduced representation learning and domain adaptation ideas for minimising the distance between control and treatment representations. TARNet, has two heads for the potential outcome regressions, while CFRNet also adds a discrepancy regulariser to TARNet for minimising the distribution distance.\\
\textbf{DragonNet, DragnonNetTR} (\cite{shi2019adapting}): They used third head for propensity score, in addition to potential outcome regressions (referred to as DragonNet), and introduced targeted regularisation (referred to as DragonNetTR).\\
\textbf{SLearner, TLearner} (\cite{kunzel2019metalearners}): These are basic baselines where the S(single)Learner augments the feature space with the treatment policy and has the advantage of using the whole data, unlike others which use data in potential outcome regression heads as per the treatment group. Similarly, T(two)Learner uses two separate networks for each treatment group and so each network utilises data related to that treatment group only.\\
\textbf{TEDVAE} (\cite{zhang2021treatment}): This uses disentangled latent factors in the variational autoencoder.\\
Moreover, the proposed disentangled representations with adversarial training algorithms are referred to as \textbf{SNet+} and \textbf{DRCFR+} due to some similarities with corresponding SNet and DRCFR algorithms. In addition, we also present extensions of DragonNet and DragonNetTR using adversarial training, referred to as \textbf{DragonNet+, DragonNetTR+}. However, adversarial training of the rest of the baselines is not possible and that's why they don't have plus-variants.

To assess the performance of the ITE estimators, since we have both factual and counterfactual outcomes in the synthetic and semi-synthetic datasets, we use Precision in the Estimation of Heterogeneous Effects (PEHE) as a metric, as given below.
\begin{equation}
    \label{eq_pehe}
    \text{PEHE} = \sqrt{\dfrac{1}{m} \sum_{i=1}^m \left(\hat{e}_i - e_i\right)^2},
\end{equation}
where $\hat{e}_i = \hat{\mu}_1^i - \hat{\mu}_0^i$ is the predicted ITE and $e_i = \mu_1^i - \mu_0^i$ is the true ITE.

\subsection{Synthetic Experiments}
\label{subsec_synthetic}
We evaluate the performance of different ITE estimators on simulated data with different sizes and with different number of confounding features. Our data generation is inspired from \cite{curth2021nonparametric}. In all cases, we take $d=25$ normally distributed covariates, subsets of which decide the potential outcomes $\mu_t(x)$ and the treatments $\pi(x)$. $X$ is generated in disjoint subsets $X_s$ of size $d_s$, according to $X_s \sim \mathcal{N}(0,1)$. Suppose, context $X$ is made of covariates $X_C, X_O, X_T, X_{\tau}$ which affect both $\mu_t(x)$ and $\pi(x)$, affect only $\mu_t(x)$, only the treatments $\pi(x)$ and model the treatment effect $e(x)$, respectively. The potential outcomes $\mu_t(x)$ and the treatments $\pi(x)$ are modelled as given below.
\begin{equation}
    \label{eq_mu0}
    \mu_0(x) = \mathbbm{1}^T X_{CO}^2,
\end{equation}
\begin{equation}
    \label{eq_mu1}
    \mu_1(x) = \mathbbm{1}^T X_{CO}^2 + \mathbbm{1}^T X_{\tau}^2,
\end{equation}
\begin{equation}
    \label{eq_pi}
    \pi(x) = expit\left(\xi \left(\dfrac{1}{d_{CT} } \mathbbm{1}^T X_{CT}^2 - \omega\right)\right),
\end{equation}
where $\mathbbm{1}$ is a vector of ones, $X_{CO} = \left[X_C,X_O\right], X_{CT} = \left[X_C,X_T\right]$, $\xi$ controls the selection-bias (in our experiments, it is set to three), and $\omega=median(\dfrac{1}{d_{CT} } \mathbbm{1}^T X_{CT}^2)$.

\begin{table}[htb!]
\centering
\caption{Comparative study of different ITE estimators on a synthetic dataset. Bold values represent the best values between the extension and the corresponding baselines and PEHE-in and -out refer to PEHE on training data and on the hold-out test dataset. The results are averaged over 10 runs, and the standard error is presented in parentheses.}
\label{tab_synthetic_results}
\begin{tabular}{llll} \\ \hline
    \textbf{Name}    & \textbf{PEHE-in} & \textbf{PEHE-out}\\ \hline
    TLearner     & 2.2822 (0.0041) & 2.5233 (0.0130)\\
    SLearner     & 0.4750 (0.0022) & 0.4772 (0.0022) \\
    TARNet       & 0.5950 (0.0041) & 0.5847 (0.0038) \\ 
    CFRNet       & 0.5950 (0.0041) & 0.5846 (0.0041) \\
    FlexTENet    & 1.7150 (0.0054) & 1.7103 (0.0060) \\ \hline
    DragonNet    & \textbf{0.5361 (0.0057)} & 0.5251 (0.0057) \\
    DragonNet+   & 0.5414 (0.0066) & \textbf{0.5220 (0.0063)} \\ \hline
    DragonNetTR  & 1.2161 (0.0145) & 1.2027 (0.0145) \\
    DragonNetTR+ & \textbf{1.0149 (0.0044)} & \textbf{0.9976 (0.0044)} \\ \hline
    DRCFR        & 0.4785 (0.0076) & 0.4745 (0.0085) \\
    DRCFR+       & \textbf{0.4742 (0.0063)} & \textbf{0.4688 (0.0076)} \\ \hline 
    SNet         & 0.4264 (0.0070) & 0.4116 (0.0063) \\
    SNet+        & \textbf{0.4147 (0.0066)} & \textbf{0.4008 (0.0073)} \\
    \hline
    \end{tabular}
\end{table}

In Table~\ref{tab_synthetic_results}, we present results on synthetic data, which have 25 covariates of which five, five and, ten covariates affect the outcome, propensity and both, and the rest model the treatment effect, respectively. The dataset has 3000 data points, 30\% of which are taken as a test set, and 30\% of the remaining are used as a validation set for hyperparameter tuning. The table shows that among the baselines, SNet and TLearner are the best and worst performers, respectively. This is expected as SNet is one of the recent techniques with a mechanism to separate confounders. On the other hand, TLearner uses two different networks without sharing information and lacks any mechanism to control confounding. Moreover, each of the two neural networks in TLearner also utilises less data as each receives data corresponding to that treatment group only. Interestingly, SLearner performs better than expected. This is likely because of sharing a single network for potential outcomes, which cancels the noise effect and can utilise the entire data for the predictions, unlike the other estimators where each potential outcome predictor utilises the data corresponding to that outcome. Although, SLearner is known to perform poorly in high-dimensional settings.

From the table, it is also clear that adversarial training improves the error in the ITE estimation (see SNet+). It is also observed that the estimators using disentanglement learning (i.e., DRCFR, SNet, and their extensions) perform comparatively better than the other baselines as they have a mechanism to separate the confounders. Moreover, shared representation-based estimators also perform better than the other, i.e., TLearner, which does not share any information between potential outcome predictions.

\begin{figure}[htb!]
    \centering
    \includegraphics[width=0.5\textwidth]{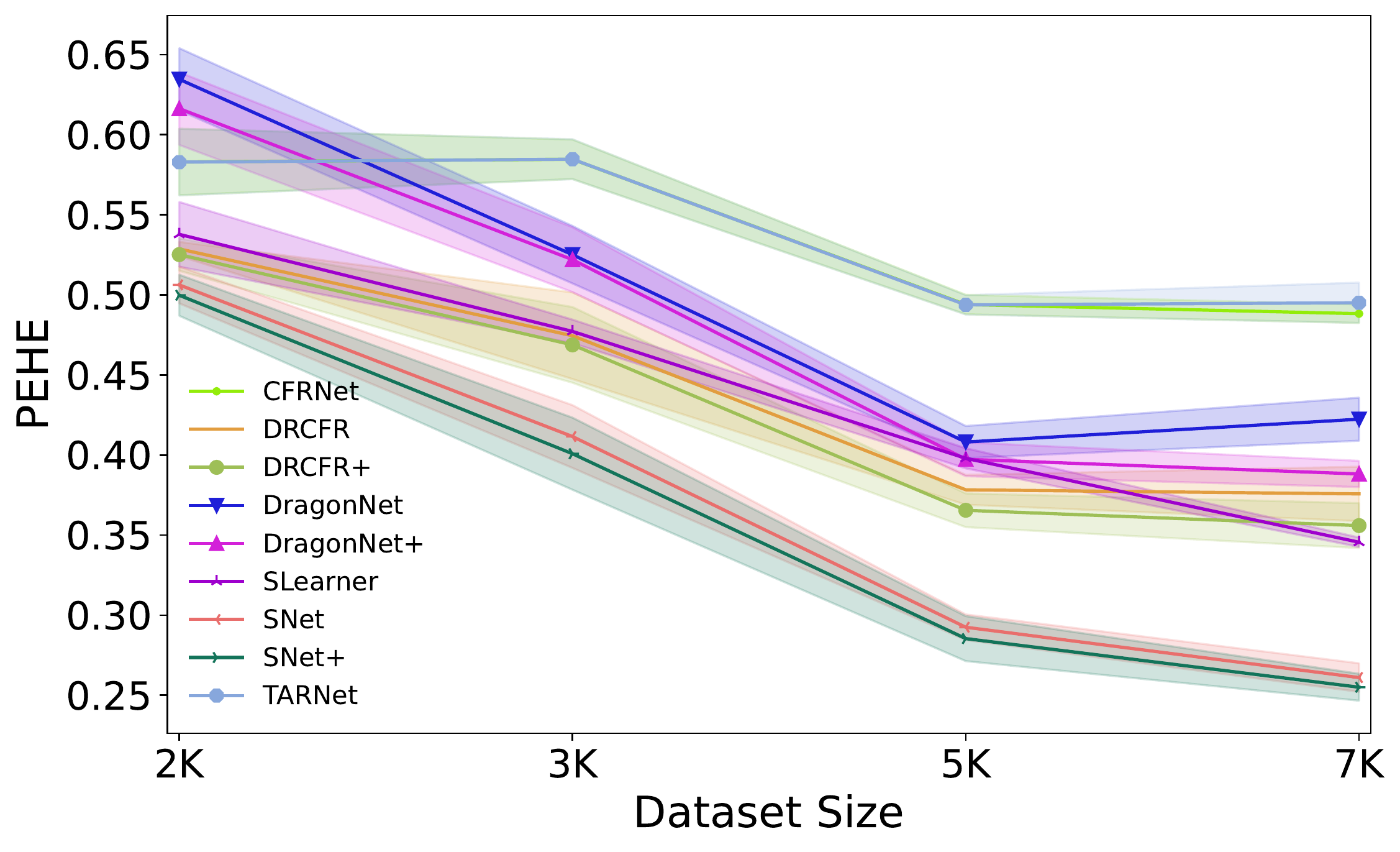}
    \caption{PEHE for different ITE estimators by dataset sizes (shaded area presents one standard deviation).}
    \label{fig_dataset}
\end{figure}

Fig.~\ref{fig_dataset} compares the performance of different ITE estimators against the scale of the data at 2,000, 3,000, 5,000, and 7,000 data points. For clarity, we removed the ITE estimators with large values, like TLearner, to avoid crowding in the figure (for a detailed comparison, please refer to the supplementary material). From the figure, it is clear that as the data size increases, in general, the error goes down. Moreover, the proposed SNet+ is always the best performer, and all the extensions are better than their baselines. It is observed that TARNet and CFRNet perform better than DragonNet and DragonNetTR with small datasets, but as the dataset size increases, they lag behind. Moreover, SNet/SNet+ improve with dataset size compared with the rest of the baselines.

\begin{figure}[htb!]
    \centering
    \includegraphics[width=0.5\textwidth]{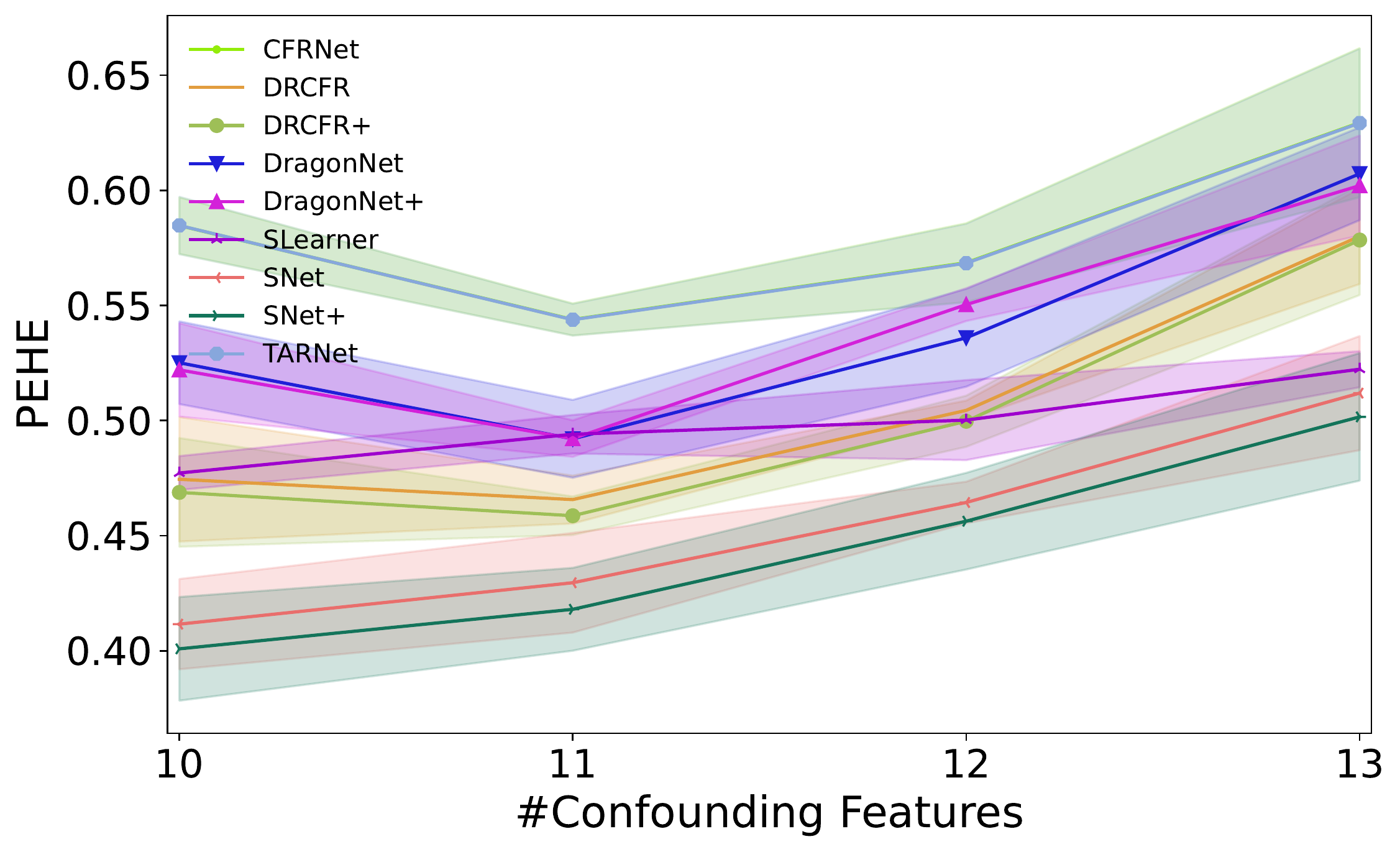}
    \caption{PEHE for different ITE estimators by different number of confounding covariates in the context $X$ (shaded area presents one standard deviation).}
    \label{fig_confounding}
\end{figure}

Fig.~\ref{fig_confounding} presents the performance of the ITE estimators for an increasing number of confounding features $\lbrace 10, 11, 12, 13 \rbrace$ with a dataset of 3000 points. All these datasets share the same covariates, i.e., $X$ is the same, although the outcomes and treatment assignment policy are calculated per the data generating process discussed earlier. From the figure, it is clear that, in general, the error increases as the number of confounding covariates increases; however, some estimators decrease at the beginning. Moreover, the proposed SNet+ and other extensions of the baselines maintain their performance and show better results. SNet+ is also the best estimator throughout and TARNet is the worst performer. As observed earlier, SLearner performs well in these settings, although it is not suited for high-dimensional settings. Once again, the disentanglement learning-based estimators perform better than other baselines.

\subsection{Semi-synthetic Benchmark: IHDP}
\label{subsec_ihdp}
The Infant Health and Development Program (IHDP) \cite{hill2011bayesian} is a semi-simulated benchmark dataset for causal inference for binary treatment problems, where the covariates are real, but outcomes are simulated. It consists of 747 units (139 treated and 608 untreated) and 25 covariates measuring aspects of mother and child. The dataset was prepared from a randomized control trial from the Infant Health and Development Program, which studies the effect of high-quality child care/specialist home visits on future cognitive test scores of the child. A biased set of treated units is removed to make the dataset imbalanced. The dataset provides noiseless true values for both the treatments, i.e., treatment and control, so it can be used to calculate the true ITE. We take IHDP100\footnote{http://www.fredjo.com/} with 100 simulation settings from \cite{shalit2017estimating} for our study. We have used a validation set as 20\% of the train set. For more details on hyperparameters, please refer to the supplementary material.

\begin{table}[htb!]
\centering
\caption{Comparative study of different ITE estimators on IHDP dataset. Bold values represent the best values between the extension and the corresponding baselines and PEHE-in, and -out refer to PEHE on training data and on hold-out test dataset. The results are averaged over 10 runs and the standard error presented in parentheses.}
\label{tab_ihdp}
\begin{tabular}{llll} \\ \hline
    \textbf{Name}    & \textbf{PEHE-in} & \textbf{PEHE-out}\\ \hline
    TLearner     & 1.1007 (0.0234) & 1.3686 (0.0211)\\			
    SLearner     & 2.5331 (0.0120) & 2.8332 (0.0101) \\ 			
    TARNet       & \textbf{0.7837 (0.0114)} & \textbf{1.0935 (0.0095)} \\ 			
    CFRNet       & 0.8165 (0.0019) & 1.1277 (0.0038) \\ 
    FlexTENet    & 1.1238 (0.0130) & 1.3954 (0.0123) \\
    TEDVAE       & 1.5004 (0.0060) & 1.6018 (0.0085) \\ \hline 		
    DragonNet    & 0.8026 (0.0054) & \textbf{1.1097 (0.0041)} \\ 
    DragonNet+   & \textbf{0.7924 (0.0104)} & 1.1125 (0.0111) \\ \hline 			
    DragonNetTR  & 1.9190 (0.0262) & 2.1935 (0.0256) \\  			
    DragonNetTR+ & \textbf{1.7087 (0.0322)} & \textbf{1.9744 (0.0338)} \\ \hline 			
    DRCFR        & 0.8680 (0.0190) & 1.1908 (0.0155) \\ 			
    DRCFR+       & \textbf{0.8616 (0.0142)} & \textbf{1.1897 (0.0095)} \\ \hline 			
    SNet         & 1.0586 (0.0066) & 1.4030 (0.0060) \\ 	 		
    SNet+        & \textbf{1.0459 (0.0082)} & \textbf{1.3839 (0.0057)} \\ 	
    \hline
    \end{tabular}
\end{table}

Table~\ref{tab_ihdp} presents experiments with the semi-simulated IHDP dataset. From the table, it is clear that all the extensions of different ITE estimators with adversarial training improve the baseline results. However, disentanglement representation-based estimators are no longer the best performer. This is in line with earlier observations \cite{curth2021nonparametric} that the simple estimators perform better than the complex estimators on the IHDP dataset, as it is a very small dataset and the \textit{overlap} assumption holds only partially. TARNet and SLearner are the best and the worst performers, respectively. 

The computational cost for the proposed approach is similar to the existing baselines, as the gradient reversal layer acts as an identity in the forward pass and only reverses the gradient during the backward pass by a factor called domain adaptation parameter. So, the proposed approach does not add any additional computational cost.

\section{Conclusion}
\label{sec_conclusion}
This paper proposes disentangled representations with adversarial training to selectively balance the confounders in the binary treatment setting. We also extended the ITE estimators having confounder representations using adversarial training to balance the confounders. Using synthetic and semi-synthetic benchmarks, we showed that the proposed idea improves the error in the ITE estimation. 
One limitation of the proposed approach is that it introduces an additional hyperparameter, called domain adaptation parameter, which has to be tuned as per the dataset to control the balancing of confounders. However, it can be tuned on the validation dataset along with other machine learning hyperparameters.
In this paper, in line with some of the literature, we studied the proposed idea in the plug-in setting for the ITE estimation. Furthermore, it would be interesting to explore this concept in the two-step meta-learners' setting.

\subsubsection*{Acknowledgements}
This work was supported in part by the National Institute for Health Research (NIHR) Oxford Biomedical Research Centre (BRC) and in part by InnoHK Project Programme 3.2: Human Intelligence and AI Integration (HIAI) for the Prediction and Intervention of CVDs: Warning System at Hong Kong Centre for Cerebro-cardiovascular Health Engineering (COCHE).
VKC was supported by a Medical Research Council (MRC) Research Grant (MR/W01761X/1). TZ was supported by an RAEng Engineering for Development Research Fellowship. TZ was supported by the RAEng Engineering for Development Research Fellowship. DAC was supported by an NIHR Research Professorship, an RAEng Research Chair, the InnoHK Hong Kong Centre for Cerebro-cardiovascular Health Engineering (COCHE), and the Pandemic Sciences Institute at the University of Oxford. 
The views expressed are those of the authors and not necessarily those of the NHS, the NIHR, the Department of Health, the InnoHK – ITC, or the University of Oxford.

% \bibliographystyle{plainnat}
% \bibliography{ML4HC}

\appendix
\section{Additional Experiments}

\subsection{Scale and Confounding Comparison for In-distributions}
\begin{figure}[htb!]
     \centering
     \begin{subfigure}[b]{0.47\textwidth}
         \centering
         \includegraphics[width=\textwidth]{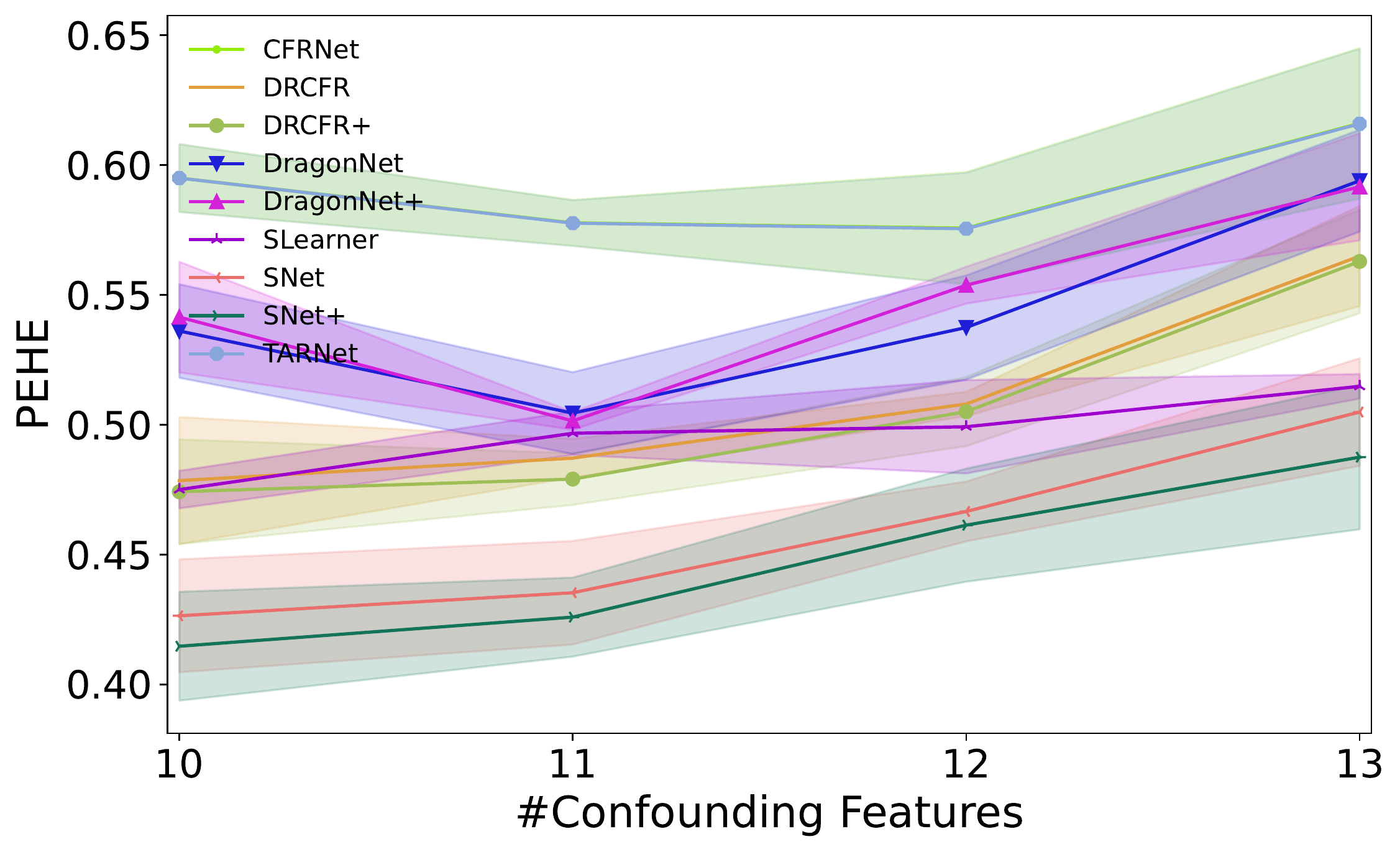}
        %  \caption{PEHE-in}
         \label{fig_confounding_in}
     \end{subfigure}
    %  \hfill
     \begin{subfigure}[b]{0.47\textwidth}
         \centering
         \includegraphics[width=\textwidth]{confounding-out.pdf}
        %  \caption{PEHE-out}
         \label{fig_confounding_out}
     \end{subfigure}
     \caption{Comparative study for in-distribution (left) and out-distribution (right) PEHE against number of confounding features.}
     \label{fig_confounders}
\end{figure}

\begin{figure}[htb!]
     \centering
     \begin{subfigure}[b]{0.47\textwidth}
         \centering
         \includegraphics[width=\textwidth]{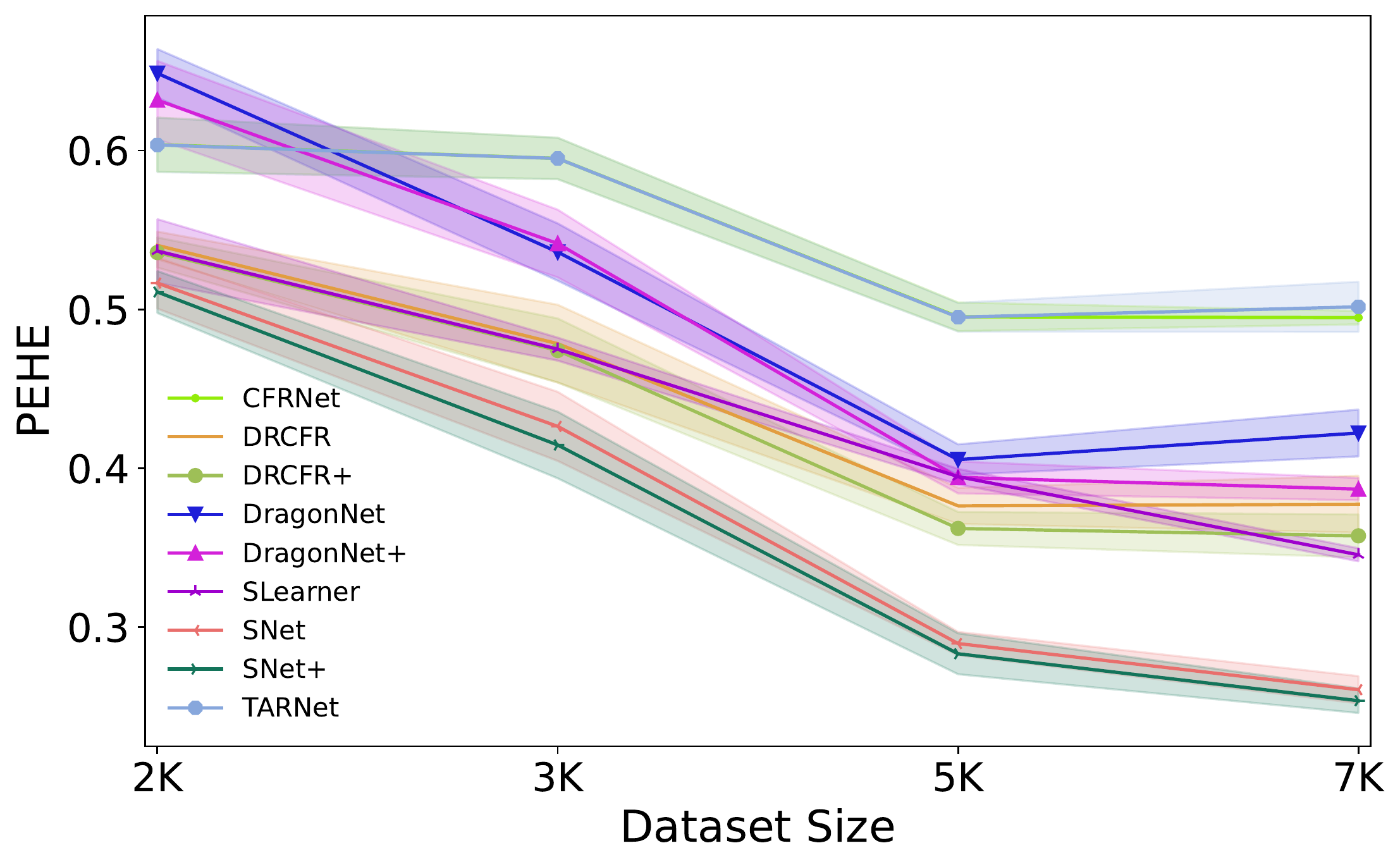}
        %  \caption{$y=x$}
         \label{fig_data_in}
     \end{subfigure}
    %  \hfill
     \begin{subfigure}[b]{0.47\textwidth}
         \centering
         \includegraphics[width=\textwidth]{dataset-out.pdf}
        %  \caption{$y=3sinx$}
         \label{fig_data_out}
     \end{subfigure}
     \caption{Comparative study for in-distribution (left) and out-distribution (right) PEHE against dataset size.}
     \label{fig_data}
\end{figure}

In the main part of the paper, we presented results for scale and confounding feature comparisons for out-distributions only. Here, we presents results for in-distributions, and for the sake of clarity, we have also presented out-distributions results side-by-side to in-distributions. From the Figs.~\ref{fig_confounders} and \ref{fig_data}, it is clear that in- and out-distribution results are similar and have same observations, as discussed earlier.

\subsection{Results with coefficients}
In the literature, either no coefficients are used to balance the potential outcome loss $L_O$ and the propensity score loss $L_T$ or coefficients are always one for the both. Although, we observed that by putting coefficients around the the two loss terms can help reduce the ITE estimation error. So, in this Subsection, we present results for IHDP dataset using the coefficients. To find the coefficients, we use manual tuning as follows.
\begin{equation}
    L = \alpha L_O + (1-\alpha) L_T + \text{other terms}.
\end{equation}

We start with a value of 0.5 for $\alpha$ and reduce/increase by 0.1 as long as it keeps improving, in the direction of improvement, and if it reaches 0.1 then to further decrease the value, we divide by 10, but if it reaches 0.9, we add extra digit 9. Table~\ref{tab_ihdp_coef} presents results on IHDP dataset with coefficients and the values of $\alpha$ are 0.6, 0.6, 0.99 and 0.99 for SNet/+, DRCFR/+, DragonNet/+ and DragonNetTR/+, respectively. For some of the baselines, there is no difference in the performance because they don't have propensity loss term so no coefficients are used for them. This is especially noted for TARNet which was earlier the best performer but with coefficients DragonNet+ performs the best. Although, the results still follow the observation from the literature that the simple ITE estimators perform better on IHDP dataset than the complex estimators. Moreover, the proposed extensions are, as observed earlier, always improve the baselines. Following the results with coefficients for IHDP dataset, we used coefficients for reporting results with synthetic data also. Although, to save the overhead to tune the coefficient, we have tuned the coefficients for setting with 3,000 data points only as they are not expected to vary much with change in data points, and values of $\alpha$ are used as 0.2, 0.5, 0.1 and 0.7 for SNet/+, DRCFR/+, DragonNet/+ and DragonNetTR/+, respectively.

\begin{table}[htb!]
\centering
\caption{Comparative study of different ITE estimators with coefficients on IHDP dataset. Bold values represent the best values between the extension and the corresponding baselines and PEHE-in, and -out refer to PEHE on training data and on hold-out test dataset. The results are averaged over 10 runs and the standard error presented in parentheses.}
\label{tab_ihdp_coef}
\begin{tabular}{llll} \\ \hline
    \textbf{Name}    & \textbf{PEHE-in} & \textbf{PEHE-out}\\ \hline
    TLearner     & 1.1007 (0.0234) & 1.3686 (0.0212)\\			
    SLearner     & 2.5331 (0.0120) & 2.8332 (0.0101) \\ 			
    TARNet       & \textbf{0.7837 (0.0114)} & \textbf{1.0935 (0.0095)} \\ 			
    CFRNet       & 0.8165 (0.0019) & 1.1277 (0.0038) \\ 
    FlexTENet    & 1.1238 (0.0130) & 1.3954 (0.0123) \\ \hline 
    DragonNet    & 0.7314 (0.0076) & 1.0563 (0.0070) \\ 
    DragonNet+   & \textbf{0.7269 (0.0092)} & \textbf{1.0478 (0.0111)} \\ \hline 			
    DragonNetTR  & 1.8887 (0.0152) & 2.1403 (0.0256) \\  			
    DragonNetTR+ & \textbf{1.6696 (0.0174)} & \textbf{1.9452 (0.0123)} \\ \hline 			
    DRCFR        & 0.8654 (0.0171) & 1.190 (0.0136) \\ 			
    DRCFR+       & \textbf{0.8418 (0.0158)} & \textbf{1.179 (0.0114)} \\ \hline 			
    SNet         & 0.99 (0.0133) & 1.30 (0.0107) \\ 	 		
    SNet+        & \textbf{0.99 (0.0044)} & \textbf{1.29 (0.0107)} \\ 	
    \hline
    \end{tabular}
\end{table}

\subsection{Comparison of Treatment Distributions}
\begin{figure}[htb!]
    \centering
    \includegraphics[width=0.5\textwidth]{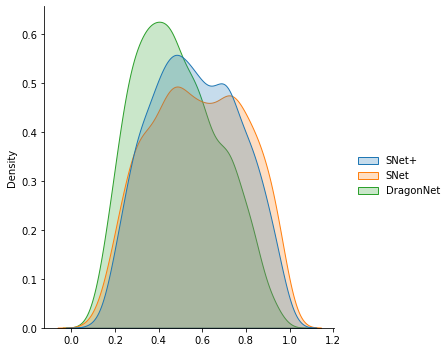}
    \caption{Distribution of treatment classification on synthetic data for DragonNet, SNet and SNet+, with PEHE-out as 0.6066, 0.5728 and 0.5677, respectively.}
    \label{fig_distr}
\end{figure}

Here, we take a random example from synthetic data, to compare the treatment distributions of SNet and SNet+, and a simple baseline of DragonNet. From the Fig.~\ref{fig_distr}, we can infer two points: first that DragonNet performs too much balancing which affects the predictive performance. This is due to the existence of a trade-off in balancing and predictive performance, and balancing the whole context $X$ as pointed out in \cite{vanderweele2019principles,kuang2019treatment}. Secondly, SNet+ improves the treatment probabilities and the density is slightly narrower than the SNet.

\section{Implementation Details}

\begin{table*}[htb!]
    \centering
    \caption{Domain adaptation parameter settings -- $\lambda_0, \gamma$ (where w5-c10-o5-3K means 5, 10 and 5 covariates determine treatment, confounders and potential outcomes with dataset size of 3000 data points. IHDP-coeff refers to IHDP dataset with coefficients for the loss terms.)}
    \label{tab_domain_param}
    \begin{tabular}{lcccc} \hline
    \textbf{Dataset} & \multicolumn{1}{l}{\textbf{SNet+}} & \multicolumn{1}{l}{\textbf{DRCFR+}} & \multicolumn{1}{l}{\textbf{DragonNet+}} & \multicolumn{1}{l}{\textbf{DragonNetTR+}} \\ \hline
    IHDP & 18, 1 & 1.2, 1 & 0.2, 600 & 0.001, 1 \\
    IHDP-coeff & 1, 1 & 1, 1 & 0.001, 1 & 0.001, 1 \\ \hline
    w5-c10-o5-2K & 1.4, 1 & 1.8, 10 & 1.5, 1 & 0.2, 600 \\
    w5-c10-o5-3K & 1.7, 1 & 1.8, 1 & 4, 1 & 1, 600 \\
    w5-c10-o5-5K & 1.7, 1 & 8, 1 & 4, 1 & 1, 600 \\
    w5-c10-o5-7K & 1.7, 1 & 15, 1 & 4, 1 & 1.5, 600 \\
    w5-c11-o5-3K & 1.7, 1 & 4, 1 & 3, 1 & 1.5, 1 \\
    w5-c12-o5-3K & 5, 400 & 1, 300 & 3, 1 & 2, 300 \\
    w5-c13-o5-3K & 0.6, 600 & 5, 1 & 3, 1 & 0.8, 600 \\ \hline
    \end{tabular}
\end{table*}
In our implementation, for fair comparison, we use components similar to \cite{curth2021nonparametric}. We set common hyperparameters for all the ITE estimators. We take a mini-batch of 100, Adam as optimiser, the learning rate of 0.0001, and $l2$-regulariser coefficient ($\lambda_1$) of 0.0001. Each experiment is run for a maximum of 1000 epochs. We use exponential linear units (ELU) as the activation function for the dense layers and early stopping based on 30\% validation (20\% for IHDP dataset due to its small size) dataset with patience of 50 as stopping criteria. For Slearner, We have used five layers with 200, 200, 200, 100 and 100 neurons, in addition to the final output layer. For TLearner, we have used five layers with 100, 100, 100, 50 and 50 neurons, in each of the two networks, in addition to the final output layers. For TARNet, CFRNet (maximum mean discrepancy), DragonNet/DragonNet+ and DragonNetTR/DragonNetTR+, we have used five layers with 200, 200 and 200 neurons in the common representation layers and 100, 100 in the potential outcome heads, in addition to the final output layer. For DRCFR/DRCFR+, we have used three representation layers with 150, 50 and 50 neurons for latent factor representations of confounder, outcome and propensity, and two layers of 100 neurons in the output heads, in addition to the final output layer. For FlexTENet, we have used two layers with 100 neurons and two layers of 50 neurons in the outcome heads, in addition to the output layer. Moreover, as suggested in the paper, we have set the $\lambda_1=0.0001, \lambda_2=0.01$ and $\lambda_o=0.1$. For SNet/SNet+, we have used 100, 100, 50, 50 and 50 neurons in the latent factor representation layers for confounder, treatment, outcome, treatment outcome and control outcome, respectively, and two layers in the outcome heads, in addition to the final output layers. We have set $\lambda_2=0$ for IHDP and $\lambda_2=0.01$ for DRCFR/DRCFR+ and SNet/SNet+. Out implementations are based on \cite{curth2021nonparametric,curth2021inductive} which are available online\footnote{https://github.com/AliciaCurth/CATENets} in JAX as well as Pytorch. For TEDVAE, we have used the official implementation and the settings from the paper (\cite{zhang2021treatment}).

The domain adaptation parameter $\lambda$ for the adversarial trained ITE estimators is also tuned manually. This requires setting the constant factor $\lambda_0$ and the parameter controlling the rate of growth $\gamma$. We vary $\gamma$ between $\lbrace 1, 10, 100, 200, 300, 400, 600, 700 \rbrace$ and $\lambda_0$ is varied between 0.1 to 25, where large values force stronger balancing of the confounder variables. The final selected values for these parameters are presented in the Table~\ref{tab_domain_param}. We observed that, the training is sensitive to $\lambda$, which has to be tuned to each dataset, because large values of $\lambda$ can lead to divergence of the learning algorithm. Since the datasets have multiple variants, e.g., IHDP has 100 so it was difficult to find a common setting for all the 100 variants because some parameters were working for variant but not for others. So, tuning of $\lambda$ for each variant on the validation dataset can give even better results at the expense of extra overhead to tune $\lambda$. Moreover, we observed that DragonNetTR+ was more sensitive to $\lambda$ and at sometimes the learning algorithm diverged so we have removed those variants from the evaluations -- including for the DragonNetTR. However, we can use gradient clipping to solve this issue of unstable learning.

\end{document}